%% file: 00_main.tex
\documentclass[10pt]{article} 
\usepackage[preprint]{tmlr}

\usepackage[utf8]{inputenc} 
\usepackage[T1]{fontenc}    
\usepackage{hyperref}       
\usepackage{url}            
\usepackage{booktabs}       
\usepackage{amsfonts}       
\usepackage{nicefrac}       
\usepackage{microtype}      
\usepackage{xcolor}         
\usepackage{amsmath}
\usepackage{pdfpages}
\usepackage{amsmath}
\usepackage{amsfonts}
\usepackage{mathtools}
\usepackage{amsthm}
\usepackage{float}

\usepackage{microtype}
\usepackage{graphicx}
\usepackage{booktabs} 

\usepackage{natbib}

\usepackage{graphicx}
\usepackage{bbm}
\usepackage{placeins}
\usepackage{adjustbox}
\usepackage{scalerel}
\usepackage[outdir=./]{epstopdf}
\usepackage{graphicx, float, pgfplots,wrapfig, sidecap, lipsum}

\usepackage{colortbl}
\usepackage{diagbox}
\usepackage{makecell}
\usepackage{tablefootnote}

\usepackage[font=small,labelfont=bf]{caption}
\usepackage{subcaption}
\usepackage{subfloat}

\usepackage[utf8]{inputenc}
\usepackage{pgfplots}
\usepgfplotslibrary{groupplots,dateplot}
\usetikzlibrary{patterns,shapes.arrows}
\pgfplotsset{compat=newest}

\usepackage{xspace}
\usepackage{soul}
\usepackage{booktabs}
\usepackage{bm}

\usepackage{listings}
\usepackage{cleveref}
 \usepackage{multirow}
\usepackage{xcolor}

\input{styles/headers}
\input{styles/math_commands}

\definecolor{keywordcolor}{rgb}{0.0, 0.0, 0.8}  
\definecolor{commentcolor}{rgb}{0.0, 0.5, 0.0}  
\definecolor{stringcolor}{rgb}{0.58, 0.0, 0.82} 
\definecolor{backgroundcolor}{rgb}{1.,1.,1.} 
\usepackage{algpseudocode}

\usepackage{hyperref}
\usepackage{url}

\title{\textit{ReFineVLA}: Multimodal Reasoning-Aware Generalist Robotic Policies via Teacher-Guided Fine-Tuning}

\author{
Tuan V. Vo$^{1}$,
Tan Q. Nguyen$^{1}$,
Khang M. Nguyen$^{2}$,
Duy H.M. Nguyen$^{3}$,
Minh N. Vu$^{4,5}$\\
\addr
$^{1}$VinRobotics, Vietnam\\
$^{2}$Max Planck Research School for Intelligent Systems (IMPRS-IS), Germany\\
$^{3}$ University of Texas at Arlington\\
$^{4}$Automation \& Control Institute, TU Wien, Austria\\
$^{5}$Austrian Institute of Technology (AIT), Vienna, Austria. Email: \texttt{minh.vu@ait.ac.at}\\
}

\begin{document}

\maketitle

\begin{abstract}
    Vision-Language-Action (VLA) models have gained much attention from the research community thanks to their strength in translating multimodal observations with linguistic instructions into robotic actions. Despite their recent advancements, VLAs often overlook the explicit reasoning and only learn the functional input-action mappings, omitting these crucial logical steps for interpretability and generalization for complex, long-horizon manipulation tasks. In this work, we propose \textit{ReFineVLA}, a multimodal reasoning-aware framework that fine-tunes VLAs with teacher-guided reasons. We first augment robotic datasets with reasoning rationales generated by an expert teacher model, guiding VLA models to learn to reason about their actions. Then, we use \textit{ReFineVLA} to fine-tune pre-trained VLAs with the reasoning-enriched datasets, while maintaining their inherent generalization abilities and boosting reasoning capabilities. In addition, we conduct an attention map visualization to analyze the alignment among visual attention, linguistic prompts, and to-be-executed actions of \textit{ReFineVLA}, showcasing its ability to focus on relevant tasks and actions. Through the latter step, we explore that \textit{ReFineVLA}-trained models exhibit a meaningful attention shift towards relevant objects, highlighting the enhanced multimodal understanding and improved generalization. 
    Evaluated across manipulation tasks, \textit{ReFineVLA} outperforms the state-of-the-art baselines. Specifically, it achieves an average increase of $5.0\%$ success rate on SimplerEnv WidowX Robot tasks, improves by an average of $8.6\%$ in variant aggregation settings, and by $1.7\%$ in visual matching settings for SimplerEnv Google Robot tasks. The source code will be publicly available. 
\end{abstract}

\input{01_introduction}

\input{02_related_work}
\input{03_preliminaries}
\input{04_method}
\input{05_experimental}
\input{06_conclusion}

\bibliographystyle{tmlr}
\bibliography{09_references}

\end{document}

%% file: styles/headers.tex
\usepackage{lipsum}  

\usepackage{hyperref}
\usepackage{url}
\usepackage{xcolor}

\usepackage{array}  

\usepackage{graphicx}  
\usepackage{listings}
\usepackage{amssymb} 
\usepackage{mathtools}
\usepackage[acronym]{glossaries}
\usepackage{siunitx,booktabs}
\usepackage[noadjust]{cite}
\usepackage{mathtools}
\usepackage{wrapfig}

\usepackage{pifont}

\usepackage{amsthm}

\usepackage[ruled,vlined,linesnumbered]{algorithm2e}
\SetCommentSty{mycommfont}
\SetKwComment{Comment}{$\triangleright$\ }{}

\crefname{assumption}{assumption}{assumptions}
\Crefname{algocf}{Algorithm}{Algorithms}

\makeatletter
\newcommand\notsotiny{\@setfontsize\notsotiny\@vipt\@viipt}
\makeatother
\makeatletter
\newcommand{\removelatexerror}{\let\@latex@error\@gobble}
\makeatother

%% file: styles/math_commands.tex
\usepackage{amsmath,amsfonts,bm}

\def\eqref#1{equation~\ref{#1}}

\def\1{\bm{1}}

\DeclareMathAlphabet{\mathsfit}{\encodingdefault}{\sfdefault}{m}{sl}
\SetMathAlphabet{\mathsfit}{bold}{\encodingdefault}{\sfdefault}{bx}{n}

\def\gD{{\mathcal{D}}}

\def\gL{{\mathcal{L}}}



%% file: 01_introduction.tex
\vspace{-10pt}
\section{Introduction}
\label{sec:intro}

Robotic manipulation policies trained on task-specific demonstrations often struggle to generalize beyond their training data, limiting their effectiveness when faced with novel objects, environments, instructions, and embodiments. Recent progress in foundation models for vision and language, such as CLIP~\citep{radford2021learning}, LLaVA~\citep{liu2024visual}, Phi-3-Vision~\citep{abdin2024phi}, along with powerful large language models (LLMs) and Vision-Language Models (VLMs), has showcased remarkable generalization across a broad spectrum of multimodality. Building on top of these, Vision-Language-Action (VLA) models~\citep{brohan2022rt, wen2024tinyvla, 3dvla, zheng2024tracevla, zhou2025chatvla, wu2024vila} has been proposed to aim for the fusion of broad generalization for foundation models with the specific expertise training on large-scale robotic datasets \citep{BerkeleyUR5Website, ebert2021bridge, zhu2023fanuc, o2023open, jiang2024robots, khazatsky2024droid}. 

Despite inheriting the robustness of VLMs and being trained on large-scale datasets, VLA models often lack sophisticated and adaptive multimodal reasoning~\citep{kim2024openvla, zhao2025cot, zheng2024tracevla, qu2025spatialvla}. They typically learn a direct functional mapping from multimodal observations to actions, without explicit step-by-step reasoning over the action horizon. This limitation becomes especially pronounced under out-of-distribution conditions, where robust performance requires a nuanced understanding of and adaptability to novel environmental variations. To address this problem, we propose \textit{ReFineVLA} to inject explicit multimodal reasoning into pre-trained VLAs. Our approach leverages an expert teacher to generate detailed natural-language rationales that articulate the sequential logic as physical intelligence behind robotic actions in the context of language-based instructions. We hypothesize that this reasoning supervision enhances the model’s understanding of observation–action relationships, thereby improving its performance on complex and compositional manipulation tasks~\citep{lu2024research, zhang2025up, zhou2025chatvla}.

Our \textit{ReFineVLA} framework fine-tunes a VLA backbone and embeds it with reasoning-augmented trajectories using a multi-objective loss to predict actions and generate multimodal reasoning rationales. By doing this, the learning of policy and explicit reasoning adaptation is jointly optimized rather than mere input–output mappings, preserving the pre-trained generalization while enhancing reasoning capacity. Specifically, we instantiate \textit{ReFineVLA} by fine-tuning a 2B-parameter VLA model on $125,000$  annotated manipulation trajectories with multimodal reasoning annotation. To evaluate generalization ability, we test our models on diverse SimplerEnv scenarios for Google and WidowX robots, which closely replicate real-world conditions. While comparing \textit{ReFineVLA}'s performance against the state-of-the-art methods, our model consistently outperforms existing VLA baselines across all embodiments and environments, demonstrating exceptional robustness under environmental variation, reflecting its deeper multimodal understanding and reasoning. To sum up, our key contributions are summarized below:
\begin{itemize}
    \vspace{-4pt}
    \item \textbf{Methodology:} We propose \textit{ReFineVLA}, a transfer-based fine-tuning framework that injects explicit multimodal reasoning into pre-trained VLAs via teacher-generated natural-language rationales, aligning policy learning with structured ``chain-of-thought" supervision.
    \vspace{-2pt} 
    \item \textbf{Dataset \& Models:} We curate a $125, 000$-trajectory dataset by prompting an expert reasoning teacher to produce step-by-step multimodal reasoning rationales for each demonstration. We fine-tune 2B VLA backbones to learn this reasoning-enriched data, leading to improved performance and inference efficiency.
    \vspace{-2pt}
    \item \textbf{Attention Visualization \& Validation:} We validate our method via extensive simulations on WidowX and Google robots across diverse embodiments. We also conduct attention-map visualizations that reveal that the model’s focus shifts from narrow action targets to semantically relevant objects and spatial anchors post–ReFineVLA fine-tuning.
    \vspace{-6pt}
\end{itemize}

%% file: 02_related_work.tex
\vspace{-0.1in}
\section{Related Work}
\label{sec:related_work}
\vspace{-0.1in}

\textbf{Vision-Language-Action Models:} Building on the success of VLMs in understanding multimodal data~\citep{karamcheti2023language, gadre2022clip, driess2023palm, du2023vision}, recent work has extended these models to robotic control, giving rise to Vision-Language-Action (VLA) models. These models aim to develop generalist robot policies by enabling pre-trained VLMs to output robot actions. Methods like RT-2~\citep{brohan2023rt}, RT-2-X~\citep{open_x_embodiment_rt_x_2023}, OpenVLA~\citep{kim2024openvla}, and RoboPoint~\citep{yuan2024robopoint} treat discretized actions as language tokens and fine-tune large VLMs on robot datasets. 
Grounding a domain-general vision–language backbone in domain-specific constraints through hybrid explicit–implicit representation learning and task-centric adaptation, as presented by Robotic-CLIP~\citep{nguyen2024robotic}, enables robots to inherit broad visual generalization while accurately modeling action-specific dynamics. Other approaches, like MobilityVLA~\citep{chiang2024mobility}, CogACT~\citep{li2024cogact}, TraceVLA~\citep{zheng2024tracevla}, and $\pi_0$\citep{black2024pi_0}, explore reasoning traces and continuous action spaces. Meanwhile, SpatialVLA\citep{qu2025spatialvla} enhances spatial representations in VLA models. Despite their strong task performance and zero-shot generalization, these models typically require complex fine-tuning for new tasks or novel robot configurations. More importantly, their action-focused training objectives often bypass explicit, step-by-step multimodal reasoning, limiting robustness in tasks that demand deeper understanding and planning~\citep{lu2024research}.

\textbf{Generalist Robot Policies:} Recent advancements in robotic learning have witnessed a significant trend towards developing multi-task ``\textit{generalist}'' robot policies capable of performing a wide array of tasks across diverse environments and embodiments, moving beyond task-specific controllers~\citep{reed2022a, brohan2022rt, haldar2023polytask, zhu2023learning}. Early efforts often focused on learning language-conditioned visual policies on a single embodiment using pretrained visual/text encoders~\citep{parisotto2015actor, rusu2015policy, shridhar2023perceiver, haldar2024baku}, which limits their adaptability to new robot platforms. More recent research leverages large-scale, cross-embodiment robot datasets~\citep{o2024open} for pre-training generalist policies, facilitating effective fine-tuning to new robot setups~\citep{team2024octo, liu2024rdt, wang2024scaling}. Notable example, Octo~\citep{team2024octo}, utilizes a flexible transformer architecture to unify embodiments. Diffusion-based generalist models, RPT~\citep{liu2024rdt} and HPT~\citep{wang2024scaling}, propose modular architectures to align data from heterogeneous embodiments. In this research line, VLAs represent a prominent direction within generalist policies by directly adapting powerful VLMs for action generation. Nevertheless, they lack the step-by-step reasoning of more sophisticated generalist robot policies. Thus, we aim to solve this problem via an expert-based fine-tuning framework that can work along with and complement generating robot policies.

\textbf{Chain-of-Thought Reasoning for Robotics:} Chain-of-Thought (CoT) reasoning has gained prominence in LLMs and VLMs, enabling them to break down complex problems into intermediate steps and improve performance on challenging reasoning tasks~\citep{wei2022chain, kojima2022large, lyu2023faithful, chia2023contrastive, yao2023beyond}. Methods like fine-tuning smaller models on rationales generated by larger ``\textit{teacher}'' models have shown promise in transferring reasoning abilities efficiently~\citep{wei2021finetuned}. CoT has also been explored in the visual domain for tasks like visual question answering and reasoning about future states~\citep{shao2024visual, rose2023visual, hu2024visual, harvey2023visual}. More recently, CoT-inspired ideas have appeared in embodied AI, including generating textual plans~\citep{mu2024embodiedgpt, michal2024robotic}, generating robotic trajectories~\citep{wen2023any, lu2023thinkbot, zawalski2024robotic}, or generating intermediate visual states~\citep{ni2024generate, liang2024dreamitate}. Nevertheless, the explicit application of multimodal reasoning to guide the policy learning process in VLA models for robotic manipulation, by training the VLA to jointly generate actions and explanatory multimodal reasoning rationales derived from a teacher, remains relatively underexplored. Our work, \textit{ReFineVLA}, also bridges this gap by explicitly leveraging multimodal reasoning rationales generated by an expert LLM teacher to guide the fine-tuning of student VLA models, thereby instilling explicit reasoning capabilities directly into the learned robotic policies.

%% file: 03_preliminaries.tex
\vspace{-0.1in}
\section{A Closer Look at Vision-Language-Action with Multimodal Reasoning}
\label{sec:prelim}
\vspace{-0.1in}

\textbf{Background:} Traditional approaches to robotic policy learning often rely on datasets of task-specific demonstrations $\mathcal{D}=\{{\tau}_1, \tau_2,...,\tau_n\}$, where each trajectory $\tau_i=\{(o_t, s_t, a_t)\}_{t=1}^T$ records expert observations, states, and actions for a single task. These methods~\citep{jain2024vcoder, zhang2025up, tschannen2025siglip} commonly employ a visual encoder $\mathcal{F}_{\phi}$ to extract features $z_i = \mathcal{F}_{\phi}(o_i)$ from image observations $o_i$, which are then fed into a policy network $\pi_{\theta}$ to output action distributions $\hat{a} \sim \pi_{\theta}(\cdot|z,s)$. Training typically involves minimizing the discrepancy between the predicted actions $\hat{a}$ and the expert actions $a$~\citep{zhang2024vla, zhen20243d, xiang2025vla}. While effective for specific tasks, this paradigm often struggles with generalization to new tasks, environments, or robot embodiments. 

\textbf{Limitations of Vision-Language-Action with Multimodal Reasoning Understanding :} Despite their great foundation in VLMs and training on large robot datasets, standard VLA models can exhibit limitations in deep multimodal understanding and reasoning since the standard VLA fine-tuning objective primarily optimizes for accurate next-action prediction, as a direct functional mapping without any presence of reasoning-based CoT~\citep{ni2024generate, liang2024dreamitate, mu2024embodiedgpt, michal2024robotic}. Given an observation $\mathbf{o}$ (which includes both image and instruction) and an expert action $\mathbf{a}$, the training process aims to minimize the negative log-likelihood of the action tokens, effectively learning the conditional probability $p(\mathbf{a} \mid \mathbf{o})$, where the standard loss formulation is $\mathcal{L}_{\text{action}} = - \log p(\mathbf{a} \mid \mathbf{o})$. Consequently, during fine-tuning, current VLA models primarily learn a direct functional mapping from visual and linguistic inputs to the corresponding action outputs. While this direct mapping is sufficient for tasks where the required action is a simple, reactive response to the immediate observation and instruction, it often fails to infuse the step-by-step multimodal reasoning needed for more complex scenarios or tasks demanding compositional physical logic~\citep{lu2024research, zhang2025up, zhou2025chatvla}. 

This restriction impacts their ability for robust multimodal understanding for reasoning tasks, such as observation-situation analysis, i.e., interpreting the current state based on visual cues, instructions \& spatial reasoning, i.e., understanding spatial relationships between objects, and the robot \& task planning, i.e., breaking down a high-level goal into sequential steps. The action prediction loss, focused solely on the final action outcome, might implicitly encourage the model to find superficial correlations sufficient for seen tasks, but without explicitly learning the underlying causal or sequential logic, leading to a deficit in robust multimodal understanding beyond direct action mapping. 

\begin{wrapfigure}{r}{0.60\textwidth}
    \centering
    \vspace{-14pt}
    \includegraphics[width=0.60\textwidth]{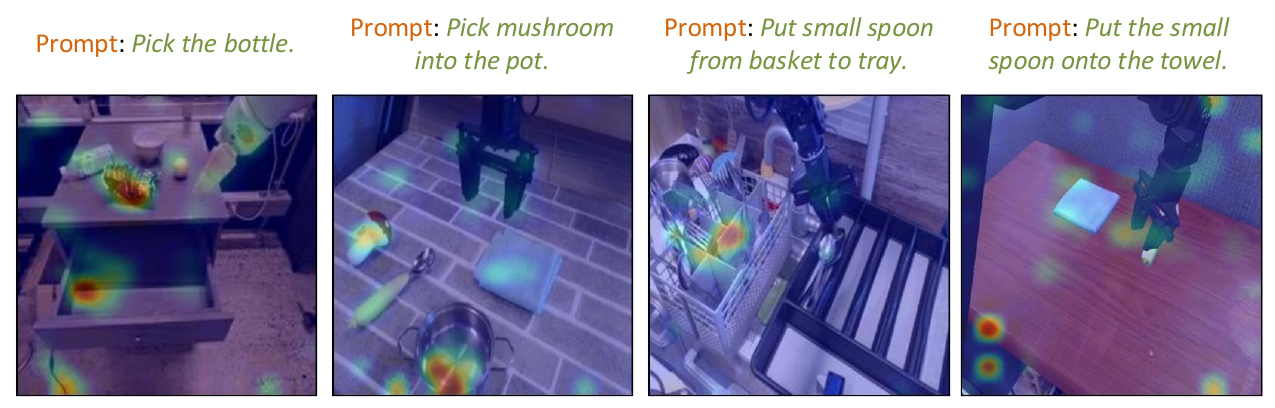}
    \vspace{-13pt}
    \caption{\textbf{Attention Visualization:} Attention maps of action tokens in standard VLAs, illustrating the narrow focus on visual cues.}
    \label{fig:visualize_attention}
    \label{fig:visualize_attention}
    \vspace{-8pt}
\end{wrapfigure}

\textbf{To investigate this limitation, we conduct a closer look at the attention mechanisms within standard VLAs}, specifically examining the attention heatmap of action tokens. Motivated by work in visual grounding~\citep{kang2025your}, which shows that certain attention heads in VLMs can localize image regions corresponding to text, we analyze where the VLA model attends in the input image when predicting actions. Our attention map visualizations in Figure~\ref{fig:visualize_attention} further reveal that traditional VLAs focus narrowly on specific visual cues directly associated with the immediate action, often disregarding crucial broader multimodal context and spatial relationships necessary for deeper reasoning. This narrow focus supports the hypothesis that standard action-only training encourages a reactive, direct mapping rather than a comprehensive multimodal understanding grounded in explicit rationale.

%% file: 04_method.tex
\vspace{-0.1in}
\section{\textit{ReFineVLA}: Reasoning-Aware Teacher-Guided Transfer Fine-Tuning}
\label{sec:method}
\vspace{-0.1in}

To enrich VLA models' capability regarding interpretability and generalization, we incorporate \textit{ReFineVLA} with an explicit multimodal reasoning. Specifically, we first present the generation of multimodal reasoning annotations in Section \ref{subsec:annotation_generation} from an expert reasoning teacher. Section \ref{subsec:selective_finetuning} outlines the selective transfer fine-tuning strategy. Thus, the training objectives are formulated in Section \ref{subsec:learning_objective} with the training algorithm (Algorithm \ref{alg:refinevla}) and the implementation details (Section \ref{subsec:implementation_details}).

\begin{figure}[t]
    \centering
    \includegraphics[width=0.88\linewidth]{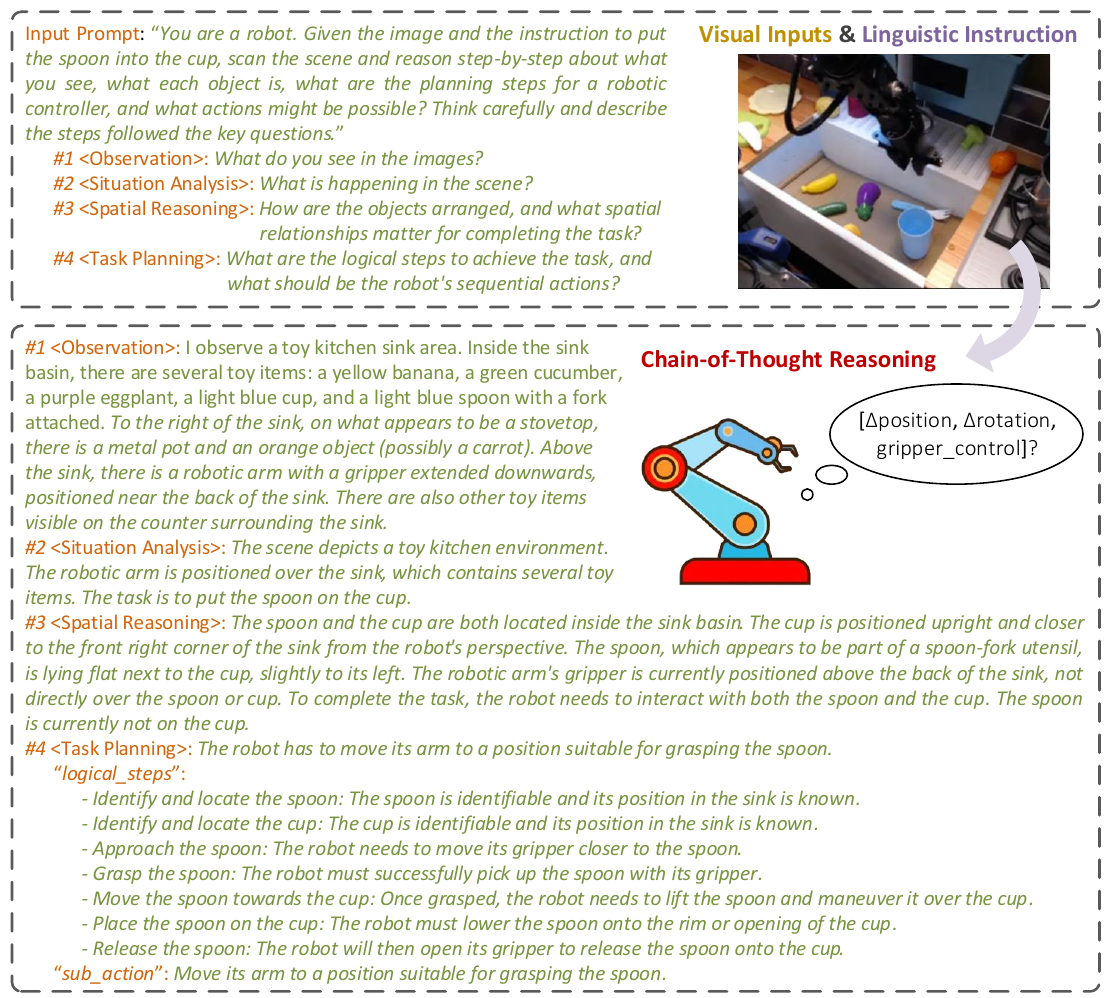}
    \caption{\textbf{Multimodal Robotic Instruction Understanding with Chain-of-Thought Reasoning}: An illustrative example depicts a single annotated data sample from a robotic manipulator grounded task planning. The task is for a robot to place a spoon into a cup, given an observation of a cluttered scene and natural language instructions. The input prompt guides the robot to reason through a sequence of structured questions: (1) \textit{Observation} -- identifying objects in the image; (2) \textit{Situation Analysis} -- understanding the context; (3) \textit{Spatial Reasoning} -- analyzing object relationships; and (4) \textit{Task Planning} -- formulating an action plan in logical steps. The annotated response includes step-by-step reasoning under each category, leading to a detailed plan of robot actions involving position, rotation, and gripper control for low-level motor commands.}
    \label{fig:cot-data-sample}
    \vspace{-18pt}
\end{figure}

\vspace{-0.1in}
\subsection{Multimodal Reasoning Annotation Generation}
\label{subsec:annotation_generation}
\vspace{-0.1in}

VLA models effectively map multimodal inputs to robotic actions, yet typically overlook explicit multimodal reasoning for interpretability and robust generalization. Standard robotic datasets, denoted as $\mathcal{D} = \left\{(o_i, a_i)\right\}_{i=1}^N$, contain multimodal observations $o_i$, including visual images $I_i$ and language instructions $l_i$, paired with actions $a_i$. These datasets generally lack explicit reasoning annotations that explain \emph{why} specific actions are suitable given the multimodal inputs.

\begin{figure}[t]
    \centering
    \includegraphics[width=0.8\linewidth]{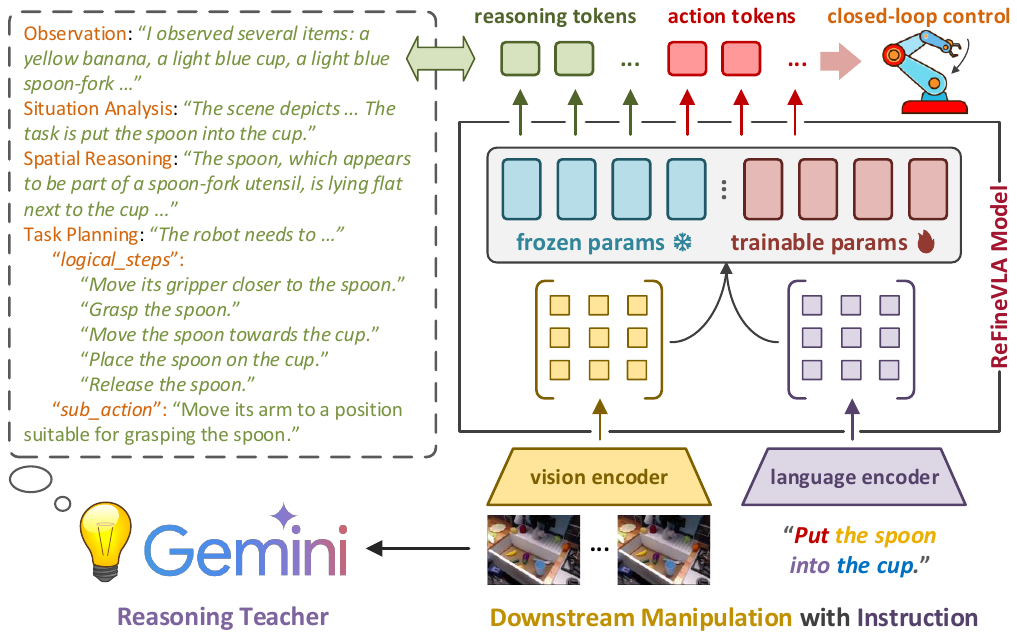}
    \caption{\textbf{\textit{ReFineVLA}'s Training Flow:} A fine-tuning framework that enhances VLA models with explicit multimodal reasoning, guided by rationales from a teacher model. These rationales cover visual cues, spatial reasoning, and task planning and are injected during training via action and reasoning losses. The learner integrates visual-linguistic inputs, infuses reasoning, and outputs interpretable actions for closed-loop control.}
    \label{fig:refinevla_overview}
    \vspace{1pt}
\end{figure}

To explicitly incorporate multimodal reasoning, we utilize powerful reasoning-based teacher models (\textit{i.e.}, Gemini \citep{team2023gemini}). We generate reasoning annotations by prompting the teacher with the structured multimodal reasoning prompt (Figure \ref{fig:cot-data-sample}). For each observation-action pair $(o_i, a_i)$, the teacher model generates a detailed multimodal reasoning annotation $r_i$, explicitly elucidating the rationale behind the chosen action given the visual and linguistic context. The enriched dataset thus becomes $\mathcal{D}' = \left\{(o_i, a_i, r_i)\right\}_{i=1}^{N}$ with $r_i$ denotes as the reasoning for $i^{\text{th}}$ pair in the original dataset $\mathcal{D}$. These annotations act as structured multimodal reasoning signals, bridging perception and action. Explicit rationales enable models to understand and reason through complex task requirements, significantly enhancing their ability to generalize across tasks and interpret their decision-making processes.

\vspace{-0.1in}
\subsection{Selective Transfer Fine-Tuning for Efficient Adaptation}
\label{subsec:selective_finetuning}
\vspace{-0.1in}

\textit{ReFineVLA} builds upon robust features learned by large-scale VLA models pretrained on extensive general robotic manipulation data. Rather than fully fine-tuning the model, which can be computationally expensive, \textit{ReFineVLA} employs selective transfer fine-tuning, outlined as follows:
\begin{itemize}
    \vspace{-4pt}
    \item \textbf{Preservation of General Features:} Pre-trained VLA models encode diverse and generalizable features. Full fine-tuning on specialized reasoning-enriched datasets risks overfitting, thereby losing foundational generalization. Therefore, selective fine-tuning preserves these foundational representations by freezing most parameters, particularly the lower layers involved in basic feature extraction.
    \vspace{-3pt}
    \item \textbf{Efficient Adaptation:} Selectively tuning a subset of model parameters significantly reduces computational resource demands, such as time, memory, and FLOPs, making \textit{ReFineVLA} practical and scalable for diverse robotic tasks and embodiments.
    \vspace{-3pt}
    \item \textbf{Targeted Knowledge Infusion:} Explicit multimodal reasoning predominantly involves higher-level cognitive processing. We hypothesize that such abstract reasoning capabilities reside in the upper layers of the VLA model, typically associated with decision-making and complex feature integration. Selective fine-tuning targets these upper layers, enabling the model to effectively integrate explicit reasoning capabilities without compromising foundational low-level feature extraction.
    \vspace{-6pt}
\end{itemize}

Empirically determined layers, such as later transformer blocks in vision and language encoders and the policy head, are selected to ensure targeted reasoning infusion, preserving general pretrained features while efficiently adapting the model for explicit reasoning.

\subsection{Learning Objective}
\label{subsec:learning_objective}
\vspace{-0.1in}

To successfully realize the shortcomings in robotic multimodal reasoning, we define the training objective of \textit{ReFineVLA} that jointly optimizes action prediction and reasoning generation as follows:
\begin{equation}
    \mathcal{L}_{\text{ReFineVLA}} = \mathcal{L}_{\text{action}} + \lambda_{\text{r}} \mathcal{L}_{\text{reasoning}},
    \label{eq:total_loss}
\end{equation}
where $\mathcal{L}_{\text{action}}$ presents the behavioral cloning loss ensuring accurate action prediction, $\mathcal{L}_{\text{reasoning}}$ guides rationale generation, fostering structured multimodal reasoning, and $\lambda_{\text{r}}$ serves as a hyperparameter controlling the penalty of \textit{ReFineVLA}'s reasoning performance.

\textbf{Action Prediction Loss ($\mathcal{L}_{\text{action}}$):} The model predicts ground-truth actions $a_i$ given multimodal inputs $o_i$, optimizing the negative log-likelihood:
\begin{equation}
    \mathcal{L}_{\text{action}} = - \sum_{t} \log \mathbb{P}(a_{i,t} \mid o_i, a_{i,<t}; \theta).
    \label{eq:action_pred_loss}
\end{equation}

\textbf{Reasoning Generation Loss ($\mathcal{L}_{\text{reasoning}}$):} This loss trains the model to produce rationales $r_i$, supervised through standard language modeling negative log-likelihood:
\begin{equation}
    \mathcal{L}_{\text{reasoning}} = - \sum_{j} \log \mathbb{P}(r_{i,j} \mid o_i, r_{i,<j}; \theta).
    \label{eq:reasoning_gen_loss}
\end{equation}
In brief. Equation \ref{eq:action_pred_loss} ensures the model learns the primary task as an objective for actions; meanwhile, Equation \ref{eq:reasoning_gen_loss} governs the model to learn thinking step-by-step, akin to a reasoning objective.

\setlength{\textfloatsep}{4pt}
\begin{algorithm}[t]
    \caption{\textit{ReFineVLA}: Reasoning-Aware Fine-Tuning of VLA}
    \label{alg:refinevla}
    \DontPrintSemicolon
    \KwIn{Reasoning dataset $\mathcal{D}' = \left\{(x_i, r_i, a_i)\right\}_{i=1}^{N}$; pre-trained parameters $\theta_{\text{full}}$; batch size $B$}
    \KwOut{Fine-tuned subset of parameters $\theta$}
    Freeze $\theta_{\text{full}} \setminus \theta$ \tcp*{only fine-tune upper layers}
    \Repeat{convergence (e.g., validation performance plateaus or training epochs reached)}{
      Sample mini-batch $\{(o_j, a_j, r_j)\}_{j=1}^B \sim \gD'$ \;
      \For{$j = 1$ \KwTo $B$}{
        $\hat{a}_j \gets \texttt{VLA}_\theta(o_j)$ \tcp*{action prediction} 
        $\hat{r}_j \gets \texttt{VLA}_\theta(o_j)$ \tcp*{auto-regressive rationale generation}
        $\mathcal{L}_{\text{action}}^{(j)} \gets -\log \mathbb{P}(a_j \mid x_j; \theta)$ \tcp*{action objective (Equation \ref{eq:action_pred_loss})}
        $\mathcal{L}_{\text{reasoning}}^{(j)} \gets -\log \mathbb{P}(r_j \mid x_j; \theta)$ \tcp*{reasoning objective (Equation \ref{eq:reasoning_gen_loss})}
      }
      $\mathcal{L}_{\text{batch}} \gets \frac{1}{B} \sum_{b=1}^{B} \left( \mathcal{L}_{\text{action}}^{(b)} + \lambda_r \mathcal{L}_{\text{reasoning}}^{(b)} \right)$ \tcp*{model objective (Equation \ref{eq:total_loss})}
      $\theta \gets \theta - \eta \nabla_\theta \gL_{\text{batch}}$ \tcp*{parameter update}
    }
\end{algorithm}

\subsection{Implementation Details}
\label{subsec:implementation_details}
The training of \textit{ReFineVLA} in Algorithm \ref{alg:refinevla} follows a supervised learning paradigm, iteratively updating the selected parameters $\theta$ of the pretrained VLA model. Through this, we are able to refine the VLA model's ability to jointly predict actions and generate rationales, leveraging pretrained knowledge while efficiently adapting to the reasoning-augmented data, as illustrated in Figure \ref{fig:refinevla_overview}.

We apply our multimodal reasoning annotation generation pipeline to the BridgeData-v2 \citep{walke2023bridgedata} and Google RT1 Robot datasets \citep{brohan2022rt} to create our datasets with reasoning annotations. Specifically, we gathered approximately $125,000$ robot trajectories annotated with multimodal reasoning annotation, forming our fine-tuning dataset for \textit{ReFineVLA}. For VLA models, we started with SpatialVLA \citep{qu2025spatialvla}, a 2B VLA model based on the PaliGemma 2 VLM backbone \citep{steiner2024paligemma}, trained on the Open X-Embodiment \citep{o2024open} and RHT20 datasets \citep{fang2024rh20t}. \textit{SpatialVLA} is also pre-trained on a large dataset, exploring effective spatial representations through techniques like Ego3D Position Encoding and Adaptive Action Grids, which enhance 3D scene understanding and transferability. These models provide robust starting points with established capabilities in generalist policies and spatial awareness, respectively, making them suitable backbones for learning enhanced reasoning.

%% file: 05_experimental.tex
\vspace{-0.1in}
\section{Experiments \& Ablation Studies}
\label{subsec:experiments}
\vspace{-0.1in}

\textbf{Baselines:} To evaluate our method, we run experiments on diverse environments, including $137$ configurations across SimplerEnv Google Robot tasks and WidowX Robot tasks that closely mimic real-world conditions. We compare against state-of-the-art open-source generalist policies with varying model sizes and training paradigms. The baselines include OpenVLA \citep{kim2024openvla}, Octo \citep{team2024octo}, RT1-X \citep{o2024open}, RoboVLM \citep{dorka2024matters}, TraceVLA \citep{zheng2024tracevla}, and SpatialVLA \citep{qu2025spatialvla}.

\subsection{Simulation Evaluation}
\textbf{SimplerEnv:} Our simulation evaluation utilizes SimplerEnv Google Robot tasks, which incorporates two distinct settings: \textit{visual matching} and \textit{variant aggregation}. The visual matching setting aims to minimize the visual appearance gap between real environments and raw simulation, significantly enhancing the correlation between policy performance in simulation and real-world scenarios. Complementing this, variant aggregation covers a wide range of environmental variations, including backgrounds of different rooms, lighter and darker lighting conditions, varying numbers of distractors, solid color and complex table textures, and camera poses. We also evaluate across different manipulation policies on the SimplerEnv WidowX Robot tasks setup with tasks: \textit{Put Spoon on Towel}, \textit{Put Carrot on Plate}, \textit{Stack Green Block on Yellow Block}, and \textit{Put Eggplant in Yellow Basket}, measuring both grasp correction success and full task completion rates. These variations allow us to assess the robustness and adaptability of \textit{ReFineVLA} in handling diverse manipulation scenarios, particularly in evaluating the spatial and temporal awareness brought by multimodal reasoning understanding.

\begin{table*}[h]
    \centering
    \vspace{-4pt}
    \caption{\textbf{Evaluated performances of different VLA baselines on SimplerEnv with WidowX Robot tasks:} The zero-shot and fine-tuning results denote the performance of the pre-trained models on the OXE dataset \citep{o2024open} and fine-tuned models on the BridgeData-v2 \citep{walke2023bridgedata}, respectively.}
    \label{tab:simplerenv_windowx}
    \resizebox{\columnwidth}{!}{
    \begin{tabular}{r cccc cccc c}
        \toprule
        \multirow{3}{*}{\diagbox{VLA Baseline}{Robotic Task}} 
        & \multicolumn{2}{c}{\makecell{Put Spoon \\on Towel}} 
        & \multicolumn{2}{c}{\makecell{Put Carrot \\on Plate}} 
        & \multicolumn{2}{c}{\makecell{Stack Green Block \\on Yellow Block}} 
        & \multicolumn{2}{c}{\makecell{Put Eggplant \\in Yellow Basket}} 
        & \multirow{3}{*}{Average} \\
        \cmidrule(lr){2-3} \cmidrule(lr){4-5} \cmidrule(lr){6-7} \cmidrule(lr){8-9}
        & Grasp & Success & Grasp & Success 
        & Grasp & Success & Grasp & Success & \\
        \midrule \midrule
        RT-1-X \citep{collaboration2023open} & 16.7\% & 0.0\% & 20.8\% & 4.2\% & 8.3\% & 0.0\% & 0.0\% & 0.0\% & 1.1\% \\
        Octo-Base \citep{team2024octo} & 34.7\% & 12.5\% & \textbf{52.8}\% & 8.3\% & 31.9\% & 0.0\% & 66.7\% & 43.1\% & 16.0\% \\
        Octo-Small \citep{team2024octo} & 
\textbf{77.8}\% & \textbf{47.2}\% & 27.8\% & 9.7\% & 40.3\% & 4.2\% & 87.5\% & 56.9\% & 30.0\% \\
        OpenVLA \citep{kim2024openvla} & 4.1\% & 0.0\% & 33.3\% & 0.0\% & 12.5\% & 0.0\% & 8.3\% & 4.1\% & 1.1\% \\
        \midrule
        RoboVLM (zero-shot) \citep{dorka2024matters} & 37.5\% & 20.8\% & 33.3\% & 25.0\% & 8.3\% & 8.3\% & 0.0\% & 0.0\% & 13.5\% \\
        RoboVLM (fine-tuning) \citep{dorka2024matters} & 54.2\% & 29.2\% & 25.0\% & 25.0\% & 45.8\% & 12.5\% & 58.3\% & 58.3\% & 31.3\% \\
        SpatialVLA (zero-shot) \citep{qu2025spatialvla} & 25.0\% & 20.8\% & 41.7\% & 20.8\% & 58.3\% & 25.0\% & 79.2\% & 70.8\% & 34.4\% \\
        SpatialVLA (fine-tuning) \citep{qu2025spatialvla} & 20.8\% & 16.7\% & 29.2\% & 25.0\% & 62.5\% & \textbf{29.2}\% & \textbf{100.0}\% & \textbf{100.0}\% & 42.7\% \\
        \midrule
        \rowcolor{gray!15} \textbf{\textit{ReFineVLA} (Ours)} & 42.9\% & 38.1\% & 33.3\% & \textbf{33.3}\% & \textbf{71.4}\% & 23.8\% & \textbf{100.0}\% & 95.2\% & \textbf{47.7}\% \\
        \bottomrule
    \end{tabular}
    }
    \vspace{-10pt}
\end{table*}

\begin{table*}[h]
    \centering
    \caption{\textbf{Evaluated performances of different VLA baselines on SimplerEnv with Google Robot tasks:} The zero-shot and fine-tuning results denote the performance of the pre-traineđ models on the OXE dataset \citep{o2024open} and the fine-tuned models on the Fractal dataset \citep{brohan2022rt}, respectively.}
    \label{tab:simplerenv_google_robot} 
    \resizebox{\columnwidth}{!}{
    \begin{tabular}{r ccc ccc}
        \toprule 
        \multirow{3}{*}{\diagbox{VLA Baseline}{Robotic Task}} & \multicolumn{3}{c}{\makecell{\textbf{Visual Matching}}} & \multicolumn{3}{c}{\makecell{\textbf{Variant Aggregation}}} \\
        \cmidrule(lr){2-4} \cmidrule(lr){5-7}
        & Move Near & \makecell{Open/Close \\Drawer} & Average & Move Near & \makecell{Open/Close \\Drawer} & Average \\
        \midrule \midrule 
        RT-1 (begin) \citep{brohan2022rt} & 5.0\% & 13.9\% & 9.5\%  & 4.0\% & 6.9\% & 5.5\% \\
        RT-1 ($15\%$) \citep{brohan2022rt} & 35.4\% & 56.5\% & 46.0\% & 44.6\% & 26.7\% & 35.7\% \\
        RT-1 (converged) \citep{brohan2022rt} & 44.2\% & \textbf{73.0}\% & 58.6\% & 50.0\% & 32.3\% & 41.2\% \\
        \midrule 
        HPT \citep{wang2024scaling} & 60.0\% & 24.0\% & 42.0\% & -- & -- & -- \\
        TraceVLA \citep{zheng2024tracevla} & 53.7\% & 57.0\% & 55.4\% & 56.4\% & 31.0\% & 43.7\% \\
        RT-1-X \citep{o2024open} & 31.7\% & 59.7\% & 45.7\% & 32.3\% & 29.4\% & 30.9\% \\
        RT-2-X \citep{o2024open} & 77.9\% & 25.0\% & 51.5\% & 79.2\% & 35.3\% & 57.3\% \\
        Octo-Base \citep{team2024octo} & 4.2\% & 22.7\% & 13.5\% & 3.1\% & 1.1\% & 2.1\% \\
        OpenVLA \citep{kim2024openvla} & 46.2\% & 35.6\% & 40.9\% & 47.7\% & 17.7\% & 32.7\% \\
        \midrule 
        RoboVLM (zero-shot) \citep{li2024towards} & 66.3\% & 26.8\% & 46.6\% & 56.0\% & 8.5\% & 32.3\% \\
        RoboVLM (fine-tuning) \citep{li2024towards} & 61.7\% & 43.5\% & 52.6\% & 60.0\% & 10.6\% & 35.3\% \\
        \midrule 
        SpatialVLA (zero-shot) \citep{qu2025spatialvla} & 69.6\% & 59.3\% & 64.5\% & 71.7\% & 36.2\% & 54.0\% \\
        SpatialVLA (fine-tuning) \citep{qu2025spatialvla} & \textbf{85.7}\% & 54.6\% & 70.1\% & 79.1\% & 26.2\% & 52.6\% \\
        \midrule
        \rowcolor{gray!15} \textbf{\textit{ReFineVLA} (Ours)} & 84.6\% & 59.0\% & \textbf{71.8}\% & \textbf{82.8}\% & \textbf{39.5}\% & \textbf{61.2}\% \\
        \bottomrule
    \end{tabular}
    }
    \vspace{-5pt}
\end{table*}

\textbf{Overall Performance:} As shown in Table~\ref{tab:simplerenv_windowx} and Table~\ref{tab:simplerenv_google_robot}, \textit{ReFineVLA} consistently outperforms state-of-the-art baselines across SimplerEnv tasks (WindowX and Google Robot). On the SimplerEnv WindowX tasks (Table~\ref{tab:simplerenv_windowx}), \textit{ReFineVLA} attains a 47.7\% average success rate, a 5.0\% improvement over SpatialVLA. Specifically, \textit{ReFineVLA} shows gains of 21.4\% on the \textit{Put Spoon on Towel} task and 8.3\% on the \textit{Put Carrot on Plate} task. For the SimplerEnv Google Robot tasks (Table~\ref{tab:simplerenv_google_robot}), \textit{ReFineVLA} excels in both settings: in variant aggregation, it achieves a 61.2\% average success rate (8.6\% over SpatialVLA); in visual matching, it reaches 71.8\% average success (1.7\% over SpatialVLA). On the \textit{Move Near} and \textit{Open/Close Drawer} tasks under variant aggregation, \textit{ReFineVLA} yields improvements of 3.7\% and 13.3\%, respectively. These results highlight \textit{ReFineVLA}’s effectiveness in incorporating explicit multimodal reasoning and enhancing generalization and robustness across diverse robotic manipulation tasks and challenging environments.

\vspace{-0.1in}
\subsection{Ablation Studies}
\label{subsec:ablation_questions}
\begin{figure}[h]
    \centering
    \vspace{-10pt}
    \includegraphics[width=1.00\linewidth]{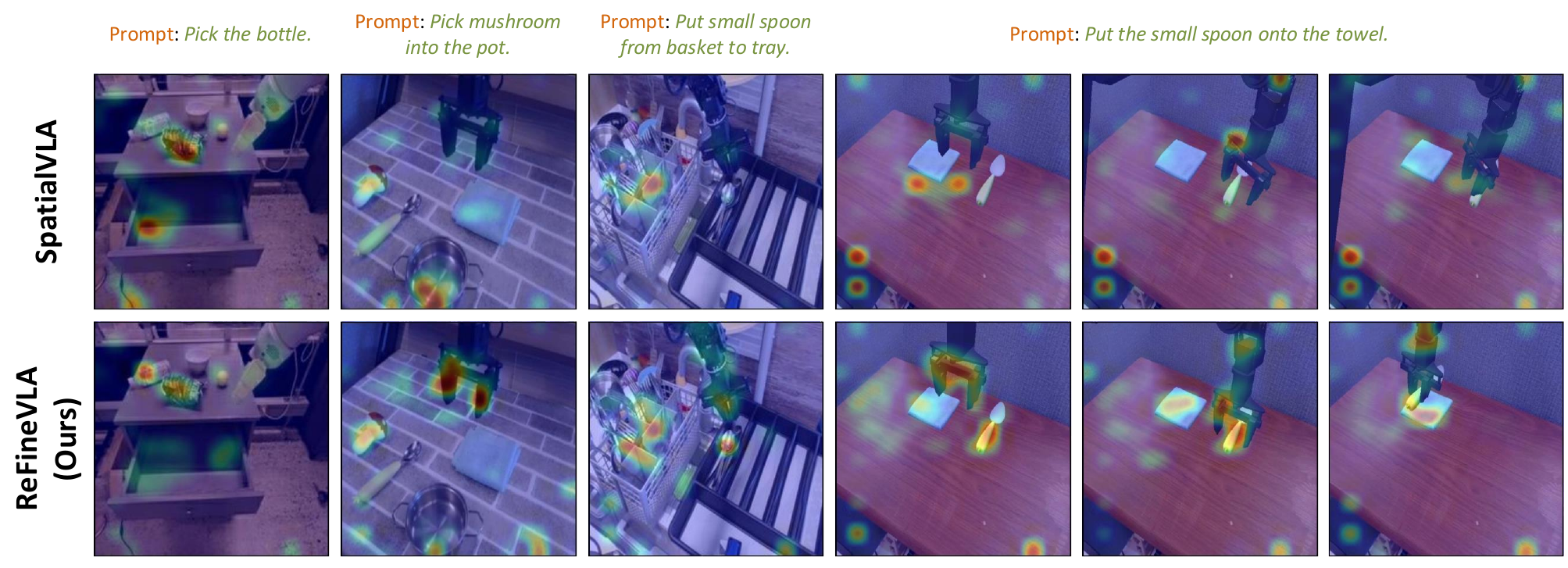}
    \vspace{-14pt}
    \caption{\textbf{Attention Visualization of \textit{ReFineVLA} compared to SpatialVLA:} \textit{RefineVLA} shows better attention to related entities within the given observations conditioned by the input prompts than what SpatialVLA does.}
    \label{fig:visualize_attention_show}  
    \vspace{-8pt}
\end{figure}

To analyze the performance gain of \textit{ReFineVLA}, we further study the following key questions:

\textbf{1) How does attention behavior change before and after \textit{ReFineVLA} fine-tuning?} To understand how reasoning supervision impacts model behavior, we analyze attention maps before and after fine-tuning with \textit{ReFineVLA}. As shown in Figure \ref{fig:visualize_attention_show}, before fine-tuning, VLA models tend to focus narrowly on immediate action targets -- often overlooking contextual elements such as the spatial arrangement of objects or instruction -- relevant cues. This behavior reflects the model's tendency to learn direct input-to-action mappings without deeper scene understanding.

After applying \textit{ReFineVLA}, we observe a consistent shift in attention toward semantically meaningful regions, empirically revealing that the model learns to reason more holistically about the task by integrating visual and linguistic information over time. Figure~\ref{fig:visualize_attention_show} shows that the model’s attention aligns better with the task instruction, supporting more robust and interpretable action decisions. These findings prove that \textit{ReFineVLA} improves performance and promotes more structured, multimodal representations and human-understandable decision-making processes.

\begin{figure}[h]
    \centering
    \vspace{-7pt}
    \includegraphics[width=1.00\linewidth]{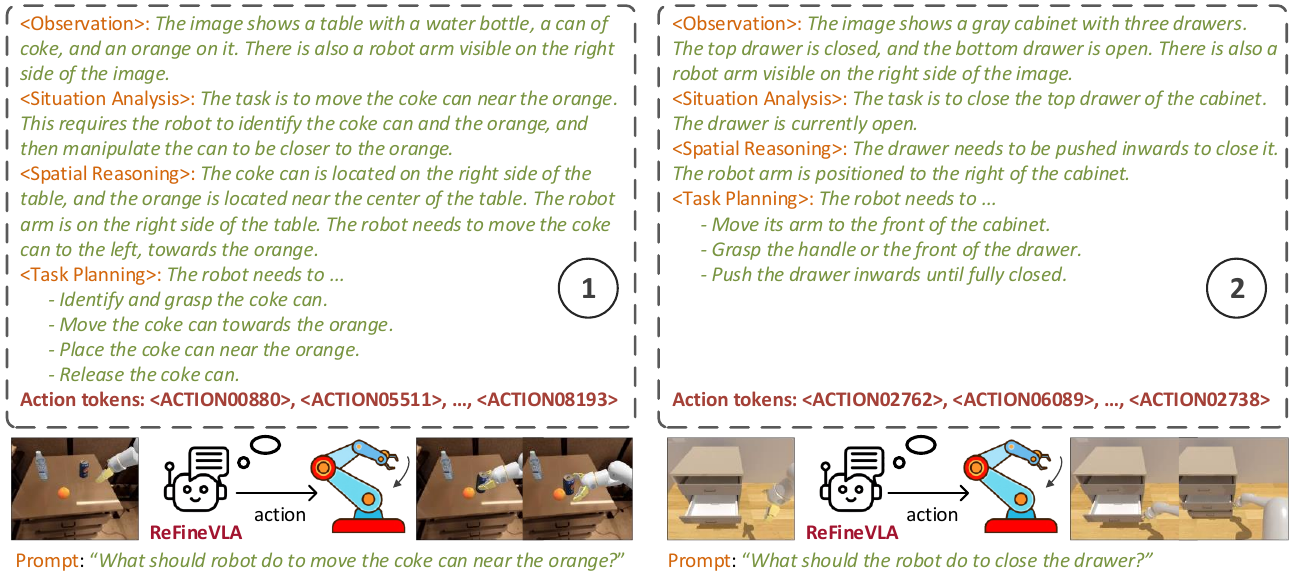}
    \vspace{-14pt}
    \caption{\textbf{Chain-of-Thought Reasoning of \textit{ReFineVLA}:} \textit{RefineVLA} shows step-by-step reasoning to accomplish the prompted task given the initial observation. The examples illustrate the queries (1) for placing the coke can near the orange positioned in the tabletop settings and (2) for closing the drawer while it is opening.}
    \vspace{-5pt}
    \label{fig:visualize_prediction}
\end{figure}

\textbf{2) How does \textit{ReFineVLA} reason on complex or long-horizon robotic manipulation tasks?} Figure \ref{fig:visualize_prediction} illustrates a reasoning procedure of \textit{ReFineVLA} for a robotic actuator actions alongside the step-by-step task planning. For complex tasks that require visual and instruction understanding, like \textit{Close the drawer}, \textit{ReFineVLA} shows a fine-grained instruction and corresponding actions to achieve the goal. \textit{ReFineVLA}'s capability continues with long-horizon tasks, such as \textit{Move the coke can near the orange}. The model can still generate proper actions with corresponding reasoning steps, not limited to observation, situation analysis, and spatial reasoning. It can be observed that \textit{ReFineVLA} can identify the objects present in embodied scenes and estimate their approximate relative positions, which is critical for action learning. Furthermore, \textit{ReFineVLA} is able to guide the robot to the next sub-actions, eventually achieving the goal task.

\begin{wrapfigure}{r}{0.5\textwidth}
    \centering
    \vspace{-13pt}
    \includegraphics[width=0.5\textwidth]{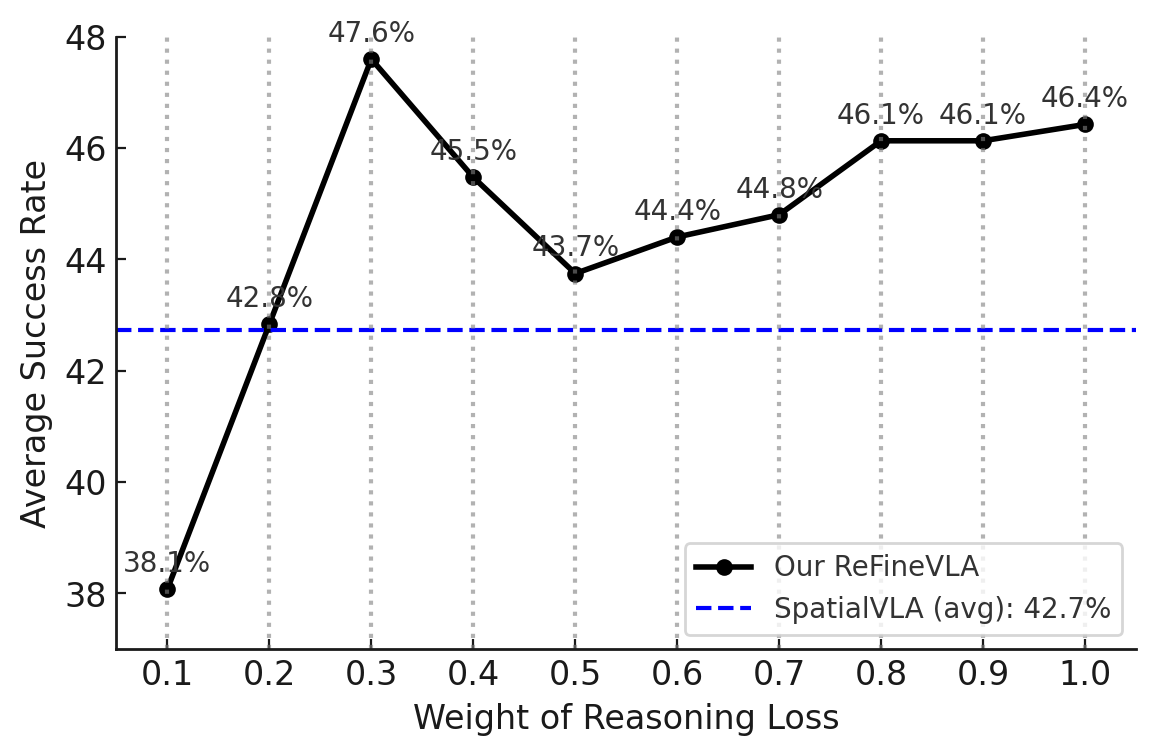}
    \vspace{-18pt}
    \caption{Average success rate of \textit{ReFineVLA} with different weights of reasoning loss $\lambda_{\text{r}}$.}
    \label{fig:avg_performance_vs_weight}
    \vspace{-10pt}
\end{wrapfigure}

\textbf{3) How does \textit{ReFineVLA} perform with different weighting of the reasoning loss, guided by the teacher's multimodal supervision?} An important hyperparameter in \textit{ReFineVLA} is the reasoning loss weight $\lambda_{\text{r}}$, which determines the strength of the reasoning loss guided by the teacher's multimodal supervision. A higher $\lambda_{\text{r}}$ encourages the model to focus more on generating detailed step-by-step rationales, potentially improving task-level understanding and interpretability. However, when $\lambda_{\text{r}}$ is set too high, it leads to overemphasis on reasoning generation, resulting in noisy outputs or misaligned attention that distracts from critical visual features such as the robot end-effector or target object. In contrast, a smaller $\lambda_{\text{r}}$ reduces the risk of distraction but may produce shallow or underdeveloped reasoning traces, weakening the benefit of teacher-guided supervision. As shown in Figure~\ref{fig:avg_performance_vs_weight}, setting $\lambda_{\text{r}} = 0.3$ achieves the best performance, yielding a $9.7\%$ improvement in task success, empirically showing that integrating teacher-provided reasoning can meaningfully enhance decision quality. However, excessively low and high values of $\lambda_{\text{r}}$ decrease performance, indicating the need to balance action prediction and reasoning learning.

\begin{wrapfigure}{r}{0.5\textwidth}
    \centering
    \vspace{-12pt}
    \includegraphics[width=0.5\textwidth]{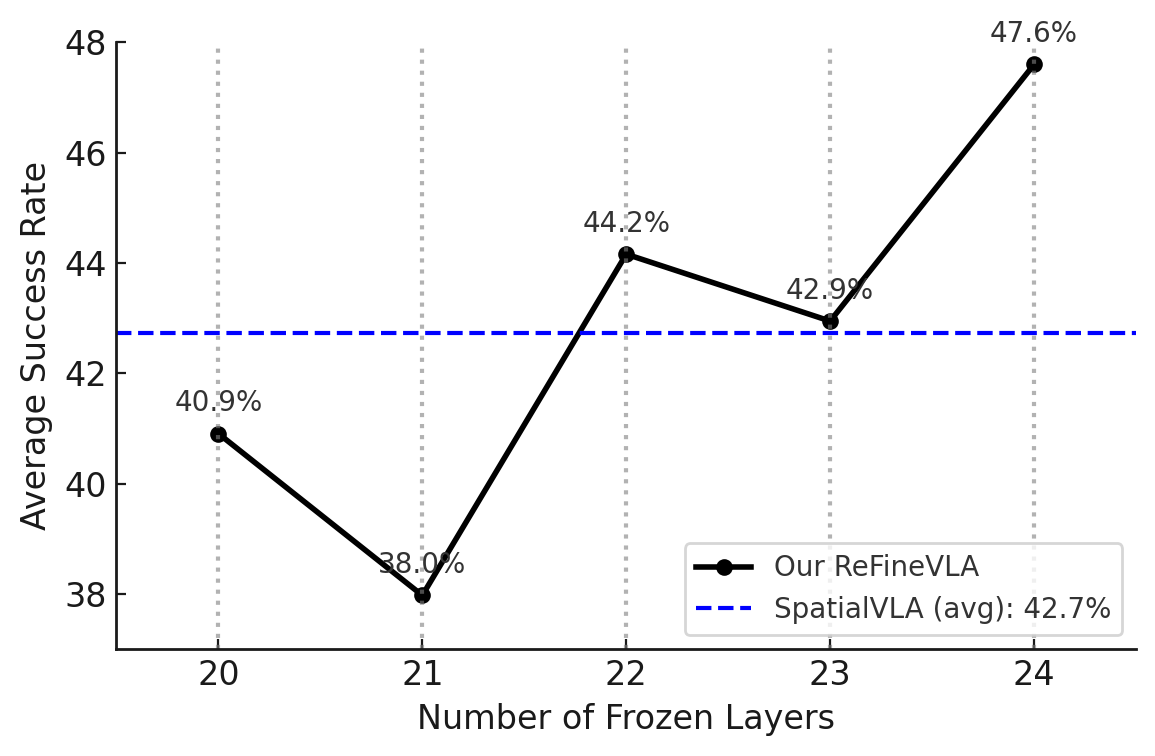}
    \vspace{-15pt}
    \caption{Average success rate of \textit{ReFineVLA} with different numbers of frozen layers during teacher-guided fine-tuning for multimodal reasoning.}
    \label{fig:avg_performance_vs_weight_freeze}
    \vspace{-17pt}
\end{wrapfigure}

\textbf{4) How does \textit{ReFineVLA} perform under different numbers of frozen layers during learning of teacher-guided multimodal reasoning supervision?}
We find that freezing lower layers helps preserve general visual and linguistic representations learned from large-scale pretraining, while allowing upper layers to specialize for structured reasoning. The performance peaks when freezing the first $24$ transformer layers, resulting in an $8.2\%$ improvement of task success rate, as shown in Figure~\ref{fig:avg_performance_vs_weight_freeze}, which suggests that the upper layers of the model are most critical for absorbing high-level reasoning supervision, while the lower layers should retain pretrained perceptual grounding.

In contrast, freezing too few layers can potentially degrade these generalizable features, leading to noisy attention and less stable reasoning outputs. Freezing too many layers restricts the model's learning capacity and limits its ability to align with teacher-guided rationales.

%% file: 06_conclusion.tex
\vspace{-0.1in}
\section{Conclusions and Discussions}
\vspace{-0.1in}
In this work, we proposed \textit{ReFineVLA}, a teacher-guided fine-tuning framework that enhances VLA models with explicit multimodal reasoning. By leveraging structured rationales from an expert reasoning teacher, \textit{ReFineVLA} trains policies that jointly predict actions and generate step-by-step reasoning, enabling more profound understanding of complex and long-horizon robotic tasks. Through selective layer tuning and penalty on weights for reasoning-governed loss, our method preserves generalizable features while injecting high-level reasoning capabilities. Moreover, experiments across simulated and real-world robotic settings in SimplerEnv with WindowX Robot and Google Robot tasks demonstrate that \textit{ReFineVLA} outperforms existing baselines in performance and interpretability, establishing a promising direction for reasoning-driven generalist VLA-based robot policies.

\textbf{Future Works}: While ReFineVLA demonstrates strong performance and improved multimodal reasoning, several promising avenues remain for exploration. One direction is to scale reasoning supervision through human-in-the-loop refinement or self-improving teacher models using reinforcement learning. Besides that, extending ReFineVLA to real-world robotic systems will also help evaluate its robustness in sim-to-real transfer settings. Finally, incorporating memory mechanisms or temporal context could enable reasoning across long-horizon tasks.